\documentclass[letterpaper, 10 pt, conference]{ieeeconf}  

\IEEEoverridecommandlockouts                              

\usepackage[T1]{fontenc}
\usepackage{graphicx} 
\usepackage{cuted}
\usepackage{capt-of}
\usepackage{epsfig} 
\usepackage{mathptmx} 
\usepackage{times} 
\usepackage{amsmath} 
\usepackage{amssymb}  
\usepackage{xcolor}
\usepackage{booktabs}

\title{\LARGE \bf
From Human Videos to Bimanual Dexterous Manipulation Data: \\ Confidence-Guided RL Refinement of Video Priors
}

\author{Yanghong Liu$^{1,*}$, Yang Chen$^{1,*}$, Kailang Xu$^{1,\ddagger}$, Bang Du$^{1,\ddagger}$, Feiyang Yang$^{1}$, Jiawei Xia$^{1}$,\\
Zhengtao Yao$^{2}$, Yubiao Zhang$^{2}$, Masayoshi Tomizuka$^{1}$, Ruihai Wu$^{1,\dagger}$, Wei Zhan$^{1}$%
\thanks{$^{*}$Yanghong Liu and Yang Chen contributed equally.
$^{\ddagger}$Kailang Xu and Bang Du contributed equally.
$^{\dagger}$Corresponding author: Ruihai Wu.}\\[0.5ex]
\authorblockA{$^{1}$University of California, Berkeley \qquad $^{2}$General Motors}
}

\begin{document}

\maketitle
\thispagestyle{empty}
\pagestyle{empty}

\begin{strip}
    \centering
    \includegraphics[width=\textwidth]{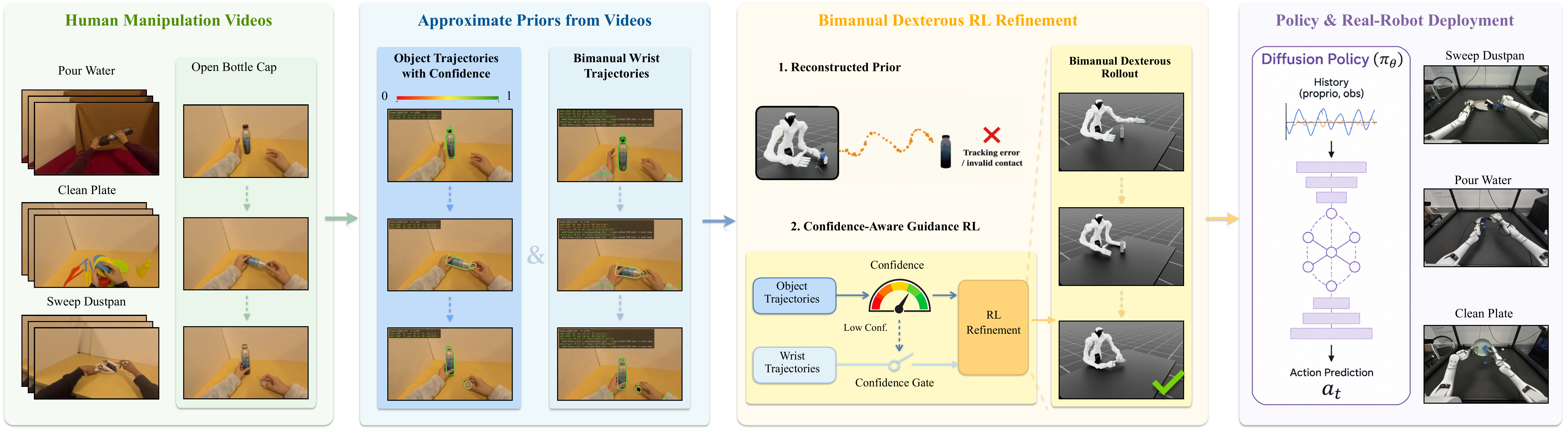}
    \captionof{figure}{
    Overview of PriDex, which transforms approximate priors extracted from human videos into executable bimanual dexterous robot demonstrations. During RL refinement, object-motion confidence adaptively balances object- and wrist-level guidance, enabling robot-feasible trajectories to be generated under physical constraints. The resulting demonstrations are then used to train a vision-based policy for downstream visuomotor control.
    }
    \label{fig:teaser}
\end{strip}

\begin{abstract}

Collecting demonstrations for bimanual dexterous manipulation is costly, motivating the use of human videos as a source of robot training data. However, uncertainty in video-derived motion estimates and differences between human and robot embodiments make direct motion transfer challenging. In this work, we introduce PriDex, a framework that transforms human-video \underline{Pri}ors into executable robot \underline{Dex}terous manipulation demonstrations through reinforcement learning in simulation.
PriDex preserves task-relevant interaction structure while allowing embodiment-specific behavior to be regenerated for the robot. Human--object interactions provide affordance cues for selecting
task-relevant, robot-specific grasps from geometry-based candidates. For motion refinement, object trajectories provide a task-level reference, while relative bimanual wrist trajectories offer complementary guidance for coordinated motion. Since object-motion estimates can become unreliable under occlusion or visual ambiguity, PriDex uses object-motion confidence to adaptively balance these two priors, emphasizing object-level guidance when confidence is high and increasing the contribution of wrist-level guidance as confidence decreases. Reinforcement learning refines these video-derived priors under the robot's kinematic and physical constraints, and the resulting policies are rolled out to generate robot demonstrations for training a vision-based diffusion policy. We evaluate PriDex on four bimanual dexterous manipulation tasks, including pouring, sweeping into a dustpan, bottle-cap unscrewing, and plate cleaning, assessing both simulation-based robot data generation and downstream visuomotor policy learning.

\end{abstract}

\section{INTRODUCTION}

Learning bimanual dexterous manipulation requires large amounts of high-quality robot interaction data. Collecting such data directly on robots is particularly demanding for high-dimensional dexterous systems, where demonstrations must coordinate two arms, multiple fingers, and contact-rich object interactions over extended horizons. Human manipulation videos provide an appealing alternative: they are abundant, naturally capture coordinated bimanual behavior, and contain rich cues about where and how objects are manipulated~\cite{hoque2026egodex,Grauman_2022_CVPR}. However, these videos do not directly provide supervision that can be readily transferred to robots. Human and robot hands differ substantially in morphology and kinematics, while object and hand motions recovered from video can be inaccurate due to occlusion, partial observability, and reconstruction uncertainty.

To extract more actionable supervision from human videos, recent approaches reconstruct 3D hand and object motions and use the resulting trajectories for retargeting, reinforcement learning, or simulation-based skill acquisition~\cite{chen2025vividex,lum2025crossing,dan2025xsim,han2026video2sim2real,paliwal2026doasido,ren2026c2dex}. However, reconstructed motions remain only approximate proxies for robot behavior. Their quality can vary due to partial observability and reconstruction uncertainty, while even accurately recovered human motions may not transfer directly to a robot because of differences in kinematics and collision constraints. Consequently, video-derived motion should not be treated as uniformly reliable or directly executable robot supervision. The key challenge is therefore not simply how faithfully to reproduce reconstructed human motion, but which cues from the video should guide robot learning and when.

Our key insight is to separate transferable interaction structure from embodiment-specific motion when transferring human demonstrations to robots. Human videos provide useful cues about where interaction occurs and how the manipulated object evolves, whereas exact human grasp configurations and hand motions are strongly embodiment-dependent and therefore need not be reproduced by the target robot. We therefore preserve interaction cues that transfer more naturally across embodiments, while allowing grasping and motion details to be regenerated and refined for the robot. This reframes human-video transfer from reproducing human motion to preserving the interaction structure most relevant to successful execution.

Following this principle, we introduce \textbf{PriDex}, a confidence-guided framework that converts human-video priors into executable bimanual dexterous robot demonstrations through reinforcement learning in simulation. PriDex first infers affordance regions from human-object interactions and uses them to sample task-relevant robot grasps from geometry-based candidates~\cite{chen2025dexonomy,chen2025bodex}, avoiding direct retargeting of human grasp configurations that may be sensitive to pose-estimation uncertainty and embodiment differences. For motion learning, object trajectories provide the primary reference for the desired evolution of the manipulated object. Since the reliability of object-motion estimates can vary over time, PriDex complements this object-level reference with wrist-motion cues that capture how each hand moves from the onset of interaction. These wrist trajectories provide a coarser motion cue that is less dependent on detailed hand morphology. PriDex then uses object-motion confidence to adaptively balance the two guidance signals, emphasizing object-level supervision when confidence is high and placing greater weight on wrist-level guidance as confidence decreases. Reinforcement learning refines these video-derived priors under the robot's kinematic and physical constraints, producing coordinated behaviors without requiring exact reproduction of the demonstrated human motion. Finally, the learned policies are rolled out in simulation to generate bimanual dexterous robot demonstrations, which are subsequently distilled into a vision-based diffusion policy~\cite{chi2023diffusionpolicy} for downstream visuomotor control.

We evaluate PriDex on four bimanual dexterous manipulation tasks from the EgoDex dataset~\cite{hoque2026egodex}, including sweeping, pouring, bottle-cap unscrewing, and plate cleaning. Starting from only three human videos per task, PriDex performs reinforcement learning in Isaac Sim~\cite{nvidiaisaacsim} to expand these demonstrations into hundreds of robot trajectories for downstream policy learning. We evaluate the resulting data-generation pipeline through simulation RL and downstream visuomotor policy learning, examining both skill acquisition and the utility of the generated demonstrations.

In summary, our contributions are as follows:
\begin{itemize}
    \item We introduce a human-video-to-robot framework that converts video-derived priors into executable bimanual dexterous robot demonstrations through reinforcement learning in simulation, allowing robot behavior to be refined beyond direct human-motion retargeting.
    \item We develop a structured prior refinement strategy that preserves transferable interaction cues while adapting embodiment-specific behavior, combining affordance-aware grasp selection with confidence-guided motion refinement.
    \item We evaluate PriDex on four bimanual dexterous tasks from EgoDex, generating thousands of robot trajectories from only 12 human manipulation videos for simulation and downstream visuomotor policy learning.
\end{itemize}

\begin{figure*}[t]
    \centering
    \includegraphics[width=\textwidth]{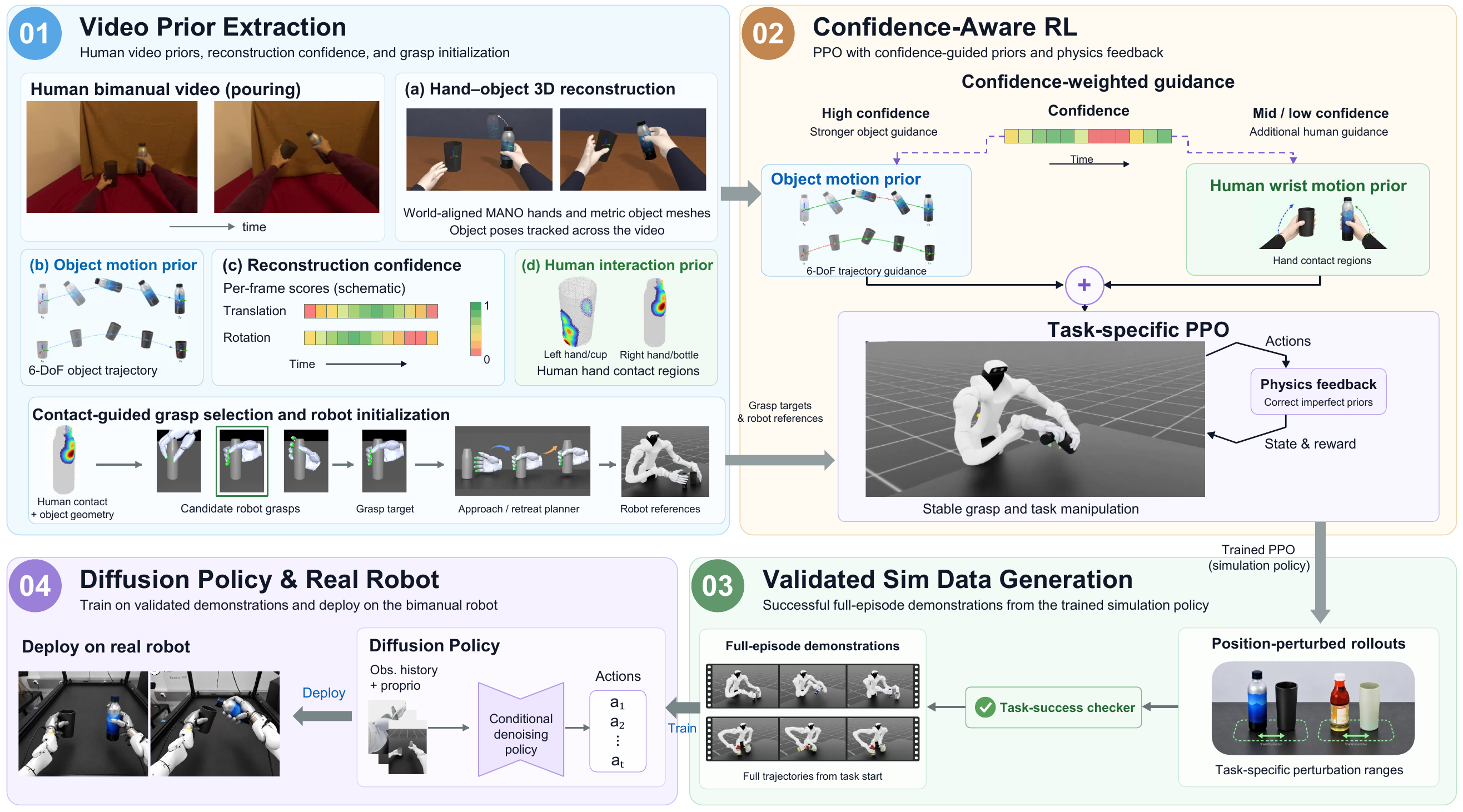}
    \vspace{-8mm}
    \caption{
    \textbf{Overview of PriDex.} Given a human manipulation video, PriDex reconstructs object motion, relative bimanual wrist motion, interaction regions, and time-varying object-motion confidence, and uses the recovered contact cues to initialize robot-specific grasps. The PPO policy then refines these priors in simulation: reliable object motion provides task-level guidance, while lower-confidence segments allow greater residual adaptation and increased wrist-motion guidance. Successful full-episode rollouts under perturbed initial conditions are retained as robot demonstrations and distilled into a vision-based Diffusion Policy for real-robot deployment.
    }
    \label{fig:pipeline}
    \vspace{-3mm}
\end{figure*}

\section{Related Work}

\subsection{Learning Robot Manipulation from Human Videos}
Human videos provide a scalable source of manipulation supervision without requiring extensive robot demonstrations. Early approaches mainly use them as high-level or embodiment-agnostic guidance. MimicPlay~\cite{wang2023mimicplay} learns latent plans from human play to guide low-level policies trained with robot demonstrations, while Im2Flow2Act~\cite{xu2025flow} uses 2D object flow as a cross-embodiment interface between human videos and simulated robot experience. More recent methods recover explicit 3D motion to provide denser supervision. X-Sim~\cite{dan2025xsim} uses object trajectories extracted from RGB-D demonstrations to guide gripper-based manipulation, whereas DexMV~\cite{qin2022dexmv}, ViViDex~\cite{chen2025vividex}, and DemoBot~\cite{xu2026demobot} rely on retargeting reconstructed human hand motion to dexterous robot hands. Recent work has begun to reduce reliance on full hand-motion imitation. Human2Sim2Robot~\cite{lum2025crossing} uses human hand poses primarily to initialize manipulation and object trajectories as task-level guidance, while DexImit~\cite{mu2026deximit} synthesizes feasible robot motion from reconstructed human demonstrations rather than directly replaying human hand trajectories. These methods progressively move beyond direct human-motion retargeting toward more selective use of video-derived supervision. PriDex instead transfers video-derived cues according to their cross-embodiment transferability and dynamically balances complementary motion priors during RL-based exploration and refinement.

\subsection{Dexterous Manipulation}
Dexterous manipulation remains challenging due to its high-dimensional action space and contact-rich dynamics, making demonstration collection considerably more demanding than for parallel-gripper manipulation. Existing systems acquire dexterous supervision through instrumented human-motion capture systems~\cite{wang2024dexcap}, direct robot teleoperation~\cite{qin2023anyteleop}, or human-to-robot motion retargeting~\cite{qin2022dexmv}. The challenge becomes more pronounced for bimanual dexterous manipulation, which requires coordinated control of two arms and multi-fingered hands. To improve data scalability, DexMimicGen~\cite{jiang2025dexmimicgen} expands a small set of source teleoperated demonstrations into large bimanual datasets through automated simulation generation. ManipTrans~\cite{li2025maniptrans} pre-trains a dexterous policy on large-scale human hand trajectories and subsequently learns task-specific residual actions for interaction-aware refinement. These approaches progressively reduce the cost of dexterous data acquisition, but still rely on costly instrumented human motion-capture systems or teleoperated robot demonstrations. PriDex instead extracts task-relevant priors from monocular human videos and uses them to generate physically executable bimanual dexterous robot rollouts through reinforcement learning.

\section{Method}
\subsection{Problem Formulation and Overview}
\label{sec:overview}
Given a human manipulation video $\mathcal{V}=\{I_k\}_{k=1}^{T}$ and the associated camera calibration, our goal is to generate bimanual dexterous robot demonstrations that preserve the task-relevant interaction in the video while adapting its realization to the robot's embodiment and physical constraints. Rather than treating reconstructed human motion as directly executable supervision, PriDex recovers structured interaction priors from the video to guide robot behavior in simulation.

As illustrated in Fig.~\ref{fig:pipeline}, PriDex first extracts object and bimanual wrist trajectories, together with time-varying object-motion confidence (Sec.~\ref{sec:priors}).
Contact-derived affordance cues guide the selection of task-relevant robot grasps from geometry-based candidates, avoiding direct transfer of human grasp configurations (Sec.~\ref{sec:grasp_init}).
These grasps initialize confidence-guided RL, with object trajectories providing task-level references and relative wrist motion offering complementary coordination guidance (Sec.~\ref{sec:rl_refinement}).
Confidence modulates these priors while RL adapts motion to robot kinematics and contact dynamics.
Finally, successful simulation rollouts train a vision-based Diffusion Policy~\cite{chi2023diffusionpolicy} for downstream visuomotor control (Sec.~\ref{sec:policy_distillation}).

\subsection{Extracting Interaction Priors from Human Videos}
\label{sec:priors}
To derive transferable motion guidance from human videos, we first reconstruct the manipulated object and recover its trajectory as the primary task-level prior. We then estimate the time-varying reliability of the recovered object trajectories and extract bimanual wrist-motion cues to provide complementary guidance for subsequent RL refinement.

\noindent\textbf{Object Geometry and Motion.}
ViPE~\cite{huang2025vipe} provides depth and camera-pose estimates, while HaWoR~\cite{zhang2025hawor} estimates wrist motion; calibrated device tracking can supply the corresponding pose inputs when available. HOI-DETR~\cite{darkhalil2026improving} initializes hand--object associations and SAM2~\cite{ravi2025sam2} mask propagation. Interaction windows identify when each hand engages its assigned object. We select reconstruction views using mask geometry, sharpness, and hand occlusion, reconstruct meshes with SAM3D~\cite{chen2026sam3d}, and estimate metric scale from depth and size cues. A customized FoundationPose-based tracker~\cite{wen2024foundationpose}, initialized near interaction onset, tracks each object forward and backward. Transforming these estimates into the world frame yields $T_k^j=(R_k^j,p_k^j)\in\mathrm{SE}(3)$ for object $j$ at video frame $k$. Inconsistent reconstructions trigger evaluation of alternative views.

\noindent\textbf{Object-Motion Confidence.}
Image alignment alone cannot establish pose reliability: occlusion limits visible evidence, tracking errors can remain locally plausible, and object symmetry can conceal rotation. We therefore combine image agreement with visibility and consistency checks, and assess rotational evidence separately. For each object at video frame $k$, translation confidence is
\begin{equation}
c_k^p=\min\{c_k^{\mathrm{img}},1-o_k,c^{\mathrm{scale}}\}\rho_k,
\label{eq:position_confidence}
\end{equation}
where the object index is omitted. Image confidence $c_k^{\mathrm{img}}$ combines visible-mask coverage and centroid alignment between the observation and projected mesh. Coverage evaluates how well the projection explains the visible object, allowing for hidden regions. Hand occlusion $o_k$ and a scale-consistency cap $c^{\mathrm{scale}}$ limit confidence when visual evidence is scarce or a silhouette-based scale correction reaches its search boundary. Taking the minimum prevents good alignment from overriding either limitation. The factor $\rho_k$ further reduces confidence for pose jumps, tracking lag, disagreement between projected point motion and CoTracker~\cite{karaev2024cotracker} tracks, and support-plane penetration, checking temporal and physical consistency beyond individual frames.

Rotation requires additional support because similar silhouettes can correspond to different orientations. We evaluate point-track consistency $\tau_k$ using CoTracker trajectories when sufficient valid tracks are available; otherwise, we use geometric observability $f_{\mathrm{rot}}$, measured by silhouette sensitivity about the least observable rotation axis:
\begin{equation}
c_k^R=
\begin{cases}
c_k^p\tau_k,&\text{sufficient valid tracks},\\
c_k^p f_{\mathrm{rot}},&\text{otherwise}.
\end{cases}
\label{eq:rotation_confidence}
\end{equation}
Point tracks can provide evidence that silhouettes lack, while the geometric fallback reduces trust in ambiguous rotations. Both factors lie in $[0,1]$, ensuring $c_k^R\leq c_k^p$.

These consistency scores control measurement weights during temporal smoothing and the strength of subsequent RL guidance.
For RL, task-specific aggregation combines the available confidence channels, followed by optional quantization and temporal alignment with the robot reference trajectory.
We denote the resulting scalar guidance signal at reference index $k$ by $c_k$.
At control step $t$, the aligned value $c_{k_t}$ jointly modulates object-tracking tolerances, arm-residual bounds, and wrist-guidance weights in Sec.~\ref{sec:rl_refinement}.
Thus, $c_{k_t}$ derives from object-motion confidence rather than an independent estimate of wrist-motion reliability.
The scores measure reconstruction consistency, not calibrated probabilities of pose correctness.

\noindent\textbf{Relative Wrist Motion.}
We preserve each hand's relative motion while allowing the robot to use its own grasp configuration. For hand $h\in\{L,R\}$, let $(\hat w_k^h,\hat R_k^h)$ denote its observed world-frame wrist pose and $u_0^h$ the first frame of its interaction window. Anchoring the demonstrated displacement and rotation at the robot grasp wrist pose $(p_0^h,R_0^h)$ from Sec.~\ref{sec:grasp_init} gives
\begin{equation}
\begin{aligned}
p_k^h&=p_0^h+\hat w_k^h-\hat w_{u_0^h}^h,\\
R_k^h&=\hat R_k^h(\hat R_{u_0^h}^h)^\top R_0^h.
\end{aligned}
\label{eq:relative_wrist}
\end{equation}
Clearance adjustment, inverse kinematics, and temporal alignment convert these wrist targets into auxiliary arm references. The construction does not prescribe finger configurations. Their induced motion directions provide complementary coordination guidance in Sec.~\ref{sec:rl_refinement} when object-motion confidence decreases.

\subsection{Affordance-Guided Robot Grasp Initialization}
\label{sec:grasp_init}

Human videos provide useful evidence about \emph{where} an object should be contacted, but the corresponding human hand configuration need not be suitable for the robot embodiment. We therefore preserve this contact structure while regenerating the grasp for the robot.  From the aligned hand--object reconstruction, we extract contact intervals and aggregate surface contact evidence in the object frame, yielding a task-conditioned affordance region $\mathcal{A}$ that captures the demonstrated interaction area.

We use $\mathcal{A}$ to guide robot-specific grasp synthesis with Dexonomy~\cite{chen2025dexonomy}. Rather than transferring the demonstrated finger configuration, Dexonomy proposes geometry-based robot grasp candidates, which we rank according to how well their intended contacts align with the video-derived affordance region. In this way, the grasp preserves the task-relevant interaction location while allowing the detailed finger configuration to adapt to the robot hand. We further refine and filter the candidates in simulation to remove configurations that violate kinematic constraints, introduce invalid collisions, or fail to establish stable contact.

Finally, a suitable hand--object grasp must also be executable by the full robot in the demonstrated scene. We therefore instantiate each retained candidate at the initial object configuration and solve for a reachable wrist pose, with cuRobo~\cite{sundaralingam2023curobo} used to obtain the corresponding arm configuration and approach motion. After the grasp closure, a brief hold test removes candidates that cannot maintain the intended contact. The validated grasp then defines the robot--object initialization for RL, allowing the subsequent stage to focus on refining the manipulation motion rather than discovering a viable grasp from scratch.

\subsection{Confidence-Guided RL Refinement}
\label{sec:rl_refinement}

The grasp initialization in Sec.~\ref{sec:grasp_init} establishes a robot-feasible contact configuration, but does not yet provide an executable manipulation trajectory. The reconstructed object motion remains only an approximate task-level prior: embodiment differences and contact dynamics may make exact tracking infeasible, while its reliability can vary over time. We therefore refine this prior with reinforcement learning in simulation. We first convert the object trajectory into a robot-specific motion reference, and then learn residual corrections whose flexibility is modulated by object-motion confidence. When confidence decreases, object tracking is relaxed and relative bimanual wrist motion provides complementary coordination guidance, allowing the robot to preserve task-relevant motion while adapting to its own physical constraints.

\noindent\textbf{Robot-Specific Motion Reference.}
To convert the object-level prior into references compatible with the robot embodiment, we propagate the robot grasps from Sec.~\ref{sec:grasp_init} along the reconstructed object trajectories. 
Let $j(h)$ denote the object associated with hand $h\in\{L,R\}$. At reference index $k$, its reconstructed pose is $\mathbf{T}^{j(h)}_k\in\mathrm{SE}(3)$, while $\mathbf{G}_h$ denotes the object-to-hand transform established at grasp initialization. The resulting robot-hand reference is
\begin{equation}
    \mathbf{T}^{\mathrm{ref}}_{H_h}(k)
    =
    \mathbf{T}^{j(h)}_k \mathbf{G}_h .
    \label{eq:hand_ref}
\end{equation}
We then solve inverse kinematics for the two hand references to obtain a nominal arm trajectory $\mathbf{q}^{\mathrm{ref}}_{\mathrm{arm}}(k)$. In this way, the reconstructed object evolution is translated into robot-specific motion without directly retargeting human hand trajectories. Since reconstruction errors and interaction dynamics may still require deviations from this nominal motion, we use $\mathbf{q}^{\mathrm{ref}}_{\mathrm{arm}}$ only as a reference for the confidence-modulated residual refinement described below.

\noindent\textbf{Confidence-Conditioned Residual Control.}
Given the nominal robot trajectory, we allow the policy to deviate from it according to the reliability of the corresponding video segment. At control step $t$, let $k_t$ denote the aligned reference frame. The PPO~\cite{schulman2017ppo} policy predicts a residual correction around the nominal reference, yielding the commanded arm target $\mathbf{q}^{\mathrm{cmd}}_{\mathrm{arm}}(t)$:
\begin{equation}
    \mathbf{q}^{\mathrm{cmd}}_{\mathrm{arm}}(t)
    =
    \mathbf{q}^{\mathrm{ref}}_{\mathrm{arm}}(k_t)
    +
    \Delta\mathbf{q}_{\mathrm{arm}}(t).
\end{equation}

As the reliability of the reconstructed object trajectory varies over time, the nominal reference should not constrain the policy uniformly. Let $c^{j(h)}_{k_t}$ denote the aligned scalar guidance signal constructed from the object-motion confidence channels in Sec.~\ref{sec:priors}. We use this confidence to determine the allowable arm-residual range $\delta(c^{j(h)}_{k_t})$:
\begin{equation}
    \left\|\Delta\mathbf{q}_{\mathrm{arm},h}(t)\right\|_\infty
    \leq \delta(c^{j(h)}_{k_t}),
    \label{eq:residual_policy}
\end{equation}
where $\delta(c^{j(h)}_{k_t})$ increases as confidence decreases. The intuition is to allocate more freedom for adaptation when the video-derived reference is less reliable: high-confidence segments use tighter object-pose tracking and smaller residual corrections, whereas lower-confidence segments relax the constraints to allow greater physics-based exploration and refinement. In practice, as confidence decreases, the position tolerance increases from $3$ to $8$\,cm and the arm-residual bound from $0.05$ to $0.10$\,rad. 

Finger articulation is handled separately from the confidence-conditioned arm refinement. While the object trajectories provide 6-DoF motion references for the hands through the nominal arm trajectory, finger-joint motion is primarily governed by the need to maintain stable contact under dexterous hand dynamics. We therefore allow bounded local adaptation around the initialized grasp: each finger joint changes by at most $1$--$2^\circ$ per control step, with its deviation from the initialized configuration capped at $35^\circ$. This preserves the overall grasp structure while allowing local contact adjustments during manipulation.

\noindent\textbf{Confidence-Adaptive Prior Balancing.}
Relaxing an unreliable object reference gives RL more freedom for physics-based exploration and refinement, but also weakens guidance for coordinated bimanual motion. We therefore introduce relative wrist motion as a complementary prior, which preserves the demonstrated coordination pattern while being less dependent on absolute human wrist poses and detailed hand morphology. Rather than matching the full wrist reference, we use only the induced arm-motion direction, retaining coarse coordination guidance while allowing the robot to adapt its absolute motion.

Let $\mathbf{v}_{\mathrm{arm}}(t)$ denote the current bimanual arm-motion direction and $\mathbf{v}^{\mathrm{wrist}}_{\mathrm{arm}}(k_t)$ the corresponding direction induced by the wrist reference. We define a directional wrist-guidance reward as
\begin{equation}
    r_{\mathrm{wrist}}(t)
    =
    \lambda(c_{k_t})
    \left[
    \cos\!\left(
        \mathbf{v}_{\mathrm{arm}}(t),
        \mathbf{v}^{\mathrm{wrist}}_{\mathrm{arm}}(k_t)
    \right)
    \right]_+,
    \label{eq:wrist_reward}
\end{equation}
where the cosine similarity measures alignment between the executed and wrist-guided motion directions, and $\lambda(c_{k_t})$ controls the guidance strength. In practice, as object-motion confidence decreases, we increase $\lambda(c_{k_t})$ from $0$ to $0.16$. Together with the confidence-conditioned object tracking above, this enables PriDex to adaptively determine how video-derived priors guide robot learning: reliable object motion provides precise task-level supervision, while lower-confidence segments increasingly rely on the more embodiment-tolerant wrist-motion cue for coordinated refinement.

\noindent\textbf{RL Objective and Training.}
The confidence-conditioned object and wrist objectives above provide the primary guidance for refining the video-derived motion. We additionally provide a terminal task-success reward upon successful task completion, together with grasp-stability and safety penalties that discourage grasp loss and joint-limit violations; for fair comparison, the same auxiliary rewards are applied to all RL baselines in our experiments.

We train PPO policies in simulation, each operating at $20$\,Hz. The policy receives state-based observations comprising robot proprioception, the current object pose and its associated confidence, and a short-horizon segment of the future reference trajectory. Since the RL policy is used only to generate robot demonstrations, it can leverage privileged simulator information that is not required by the final visuomotor policy at deployment.

\subsection{Policy Distillation}
\label{sec:policy_distillation}

We distill the privileged RL teachers into a vision-based
  Diffusion Policy~\cite{chi2023diffusionpolicy} through supervised
  imitation. Deterministic teacher rollouts are collected under
  randomized initial conditions, and only successful trajectories are
  retained to form $\mathcal{D}_{\mathrm{sim}}$. A student observation
  $\mathbf{o}_t$ consists of multi-view images together with the
  proprioceptive state $\mathbf{q}_t$, comprising all joint positions
  and both wrist poses; privileged simulator states and
  reference-trajectory information are excluded. Each action
  $\mathbf{a}_t\in\mathbb{R}^{58}$ specifies absolute joint position
  targets for the full 58-DoF bimanual system---two 7-DoF arms and
  two 22-DoF hands---commanded at 20~Hz. Whereas the teacher emits
  bounded residuals about a nominal reference, we record the composed
  targets actually sent to the controller; the student therefore
  predicts joint commands directly and requires no reference
  trajectory at inference.

\noindent\textbf{Conditional action diffusion.}            
  Given $H_o=2$ consecutive observations,
  $\mathbf{O}_t=(\mathbf{o}_{t-1},\mathbf{o}_t)$, the policy predicts
  an action chunk
  $\mathbf{A}_t=(\mathbf{a}_t,\ldots,\mathbf{a}_{t+H_p-1})$
  with $H_p=16$. We model the conditional action distribution
  using a denoising diffusion probabilistic model (DDPM)~\cite{ho2020denoising}.
  During training, a normalized action chunk
  $\bar{\mathbf{A}}_t$ is perturbed with Gaussian noise:
  \begin{equation}
  \label{eq:forward_diffusion}
  \mathbf{A}_t^{(k)}
  =
  \sqrt{\bar{\alpha}_k}\,\bar{\mathbf{A}}_t
  +
  \sqrt{1-\bar{\alpha}_k}\,\boldsymbol{\epsilon},
  \end{equation}
  where $\boldsymbol{\epsilon}\sim\mathcal{N}(\mathbf{0},\mathbf{I})$,
  $\bar{\alpha}_k=\prod_{j=1}^{k}(1-\beta_j)$, and $\beta_j$ follows
  a squared-cosine schedule. The observation-conditioned
  denoiser $\boldsymbol{\epsilon}_\theta$ learns to predict the
  injected noise by minimizing
  \begin{equation}
  \label{eq:diffusion_loss}
  \mathcal{L}_{\mathrm{diff}}
  =
  \mathbb{E}
  \left[
  \left\|
  \boldsymbol{\epsilon}
  -
  \boldsymbol{\epsilon}_\theta
  (\mathbf{A}_t^{(k)},k,\mathbf{O}_t)
  \right\|_2^2
  \right],
  \end{equation}
  where the expectation is over observation--action chunks sampled
  from $\mathcal{D}_{\mathrm{sim}}$, Gaussian noise
  $\boldsymbol{\epsilon}$, and diffusion timesteps
  $k\sim\mathrm{Uniform}(\{1,\ldots,K\})$, with $K=100$.
  The same $K$ denoising steps are used at inference to sample an action chunk.

\noindent\textbf{Architecture and training.}
  Each camera view is encoded by a separate ResNet-18 trained
  from scratch. Visual features and the proprioceptive state over
  the observation window are concatenated into a global conditioning
  vector, which modulates a one-dimensional convolutional U-Net
  denoiser via FiLM. Actions and proprioceptive inputs are normalized
  per dimension, and random image crops are used for augmentation.
  For each task, $\mathcal{D}_{\mathrm{sim}}$ pools the successful
  rollouts of the three per-video RL teachers, $200$ episodes per
  video, and a single policy is trained on it for $8$ epochs using
  AdamW with an initial learning rate of $10^{-4}$, a cosine
  learning-rate schedule, and a batch size of $32$. An exponential
  moving average of the model weights is maintained for inference.

\section{Experiments}

\subsection{Experimental Setup}
\begin{figure}[t]
    \centering
    \includegraphics[width=\linewidth]{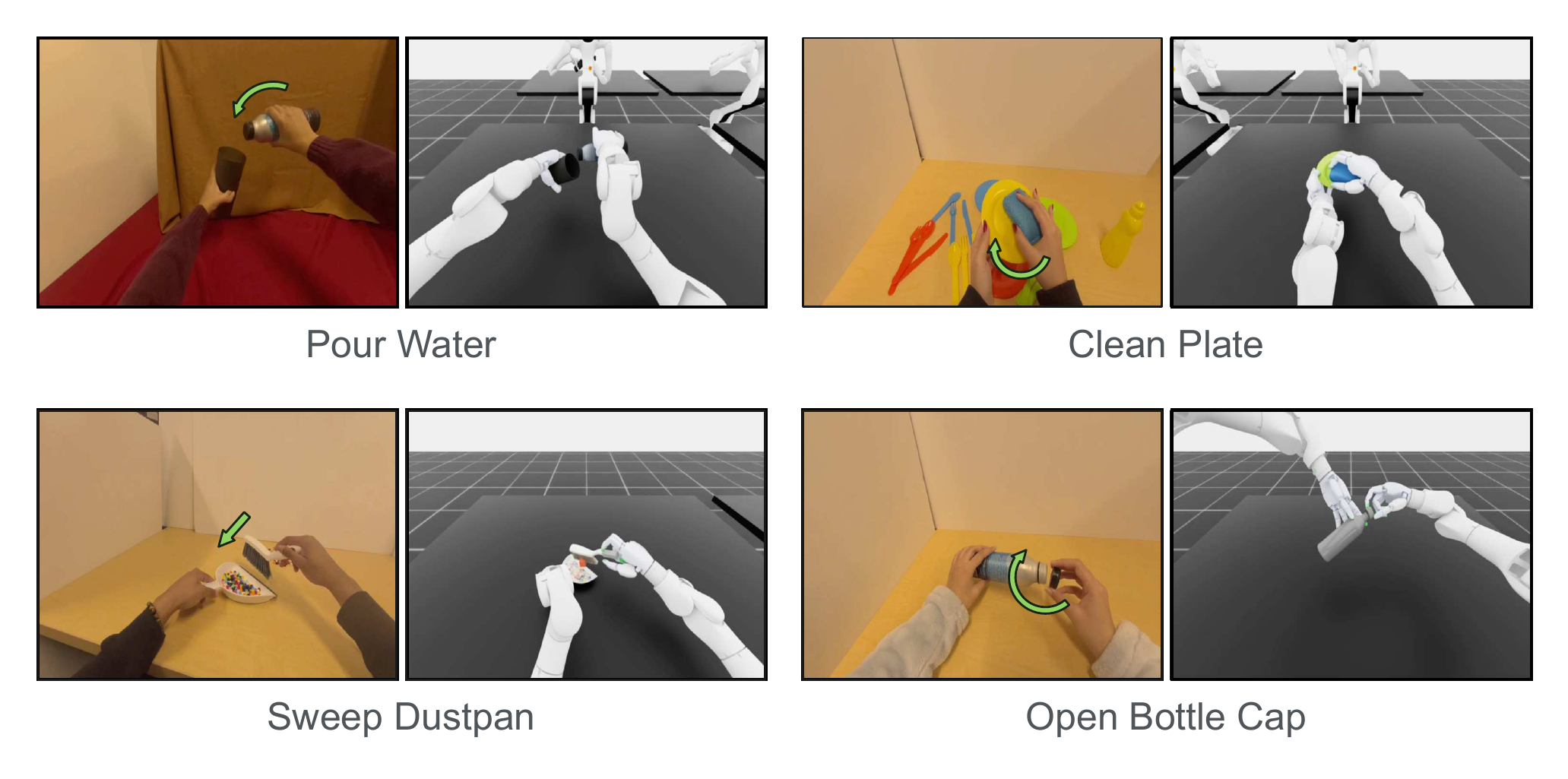}
    \vspace{-8mm}
    \caption{Four evaluated bimanual manipulation tasks. Each pair shows a human demonstration (left) and the corresponding simulated robot task (right): pouring, plate cleaning, sweeping, and bottle-cap unscrewing.}
    \label{fig:four_tasks}
    \vspace{-2mm}
\end{figure}

We conduct simulation experiments in Isaac Sim 5.1 using a Dexmate bimanual robot equipped with two Sharpa dexterous hands.
We evaluate four tasks shown in Fig.~\ref{fig:four_tasks}: pouring (Pour), sweeping into a dustpan (Sweep), bottle-cap unscrewing (Unscrew), and plate cleaning (Clean).
Each task uses three source videos, and each evaluated RL configuration receives a training budget of 20 million environment steps per source video.

For both RL and Diffusion Policy evaluation, we perturb task-relevant initial positions in the tabletop plane, sampling each coordinate offset from $[-a,a]$. We set $a=3$\,cm for Pour, $a=1$\,cm for Clean, and $a=5$\,mm for Sweep and Unscrew. In Pour, Unscrew, and Clean, the manipulated objects undergo a common translation that preserves their relative configuration. In Sweep, we perturb the debris position.

For policy distillation, we collect separate datasets of successful rollouts from full PriDex and its variant without human-motion guidance, using the same position perturbations. Each dataset contains 200 demonstrations per source video, totaling 600 per task. We train one task-specific Diffusion Policy~\cite{chi2023diffusionpolicy} per data source with matched demonstration counts. Each policy takes RGB images from one head-mounted and two wrist-mounted cameras, together with robot proprioception, and predicts absolute joint-position targets.

\subsection{Performance and Analysis of PriDex}

\subsubsection{Baselines}
We compare PriDex with task-adapted implementations of
Human2Sim2Robot (H2S2R)~\cite{lum2025crossing} and
ViViDex~\cite{chen2025vividex}. H2S2R uses reconstructed object
trajectories as dense tracking objectives and initializes exploration
from a retargeted pre-manipulation pose. We adapt its single-arm policy
interface by duplicating and concatenating the observation and action
spaces for bimanual control. Unlike PriDex, H2S2R relies on a fixed
object-tracking objective rather than confidence-conditioned object- and
wrist-motion guidance.

ViViDex trains state-based PPO teachers using reconstructed human
hand--object trajectories. We extend its single-hand control and trajectory
targets to paired left- and right-hand channels for bimanual execution.
Unlike PriDex, ViViDex directly treats the reconstructed trajectories as
policy guidance without confidence-based adaptation or affordance-guided
robot grasp generation.

All methods use the same RL interaction budget.

\subsubsection{Ablations}
We design two cumulative ablations to examine the roles of confidence in RL refinement: auxiliary motion guidance and adaptive exploration.
Object trajectories describe the intended task behavior, whereas human hand trajectories provide one possible means of achieving it.
Relative wrist motion therefore serves as auxiliary guidance when object-motion references are unreliable.
At low confidence, our method strengthens wrist guidance and expands exploration through two independent mechanisms: increasing per-step and cumulative residual-action bounds, and relaxing positional and rotational tolerances in reference-tracking penalties.

\emph{w/o Human Prior (HP)} removes the relative wrist-motion guidance in Eq.~\eqref{eq:wrist_reward} while retaining confidence-adaptive action bounds and tracking tolerances.
Comparison with the full method tests whether wrist guidance improves task completion when adaptive exploration remains available.
Affordance-guided grasp initialization and object-motion references are retained; HP thus denotes only the relative wrist-motion prior.

\emph{w/o HP \& Conf.} additionally disables confidence adaptation throughout refinement, holding the per-step and cumulative residual-action bounds fixed and setting the positional and rotational tracking tolerances to $5$\,cm and $30^\circ$, respectively.
Comparison with \emph{w/o HP} tests whether confidence-adaptive refinement improves performance over fixed settings in the absence of wrist guidance.

\subsubsection{Evaluation Metrics}
\label{sec:metrics}
We report task success rate as the percentage of evaluation episodes
satisfying the task-specific completion criterion. For each task,
we evaluate the policies trained from the three source videos for
$50$ episodes each, yielding $150$ evaluation episodes in total.
Task success rates are computed over these pooled episodes.
Mean success is the unweighted average across the four tasks.

\begin{table}[htbp]
    \centering
    \caption{Simulation RL success rates (\%). Best per task in bold.}
    \label{tab:sim_rl_results}
    \footnotesize
    \setlength{\tabcolsep}{2.3pt}
    \renewcommand{\arraystretch}{1.08}
    \begin{tabular*}{\linewidth}{@{\extracolsep{\fill}}lccccc@{}}
        \toprule
        Method & Pour & Sweep & Unscrew & Clean & Mean \\
        \midrule
        H2S2R
        & 0.0 & 3.0 & 0.0 & 7.3 & 2.6 \\
        ViViDex
        & 0.0 & 6.0 & 0.0 & 0.0 & 1.5 \\
        \midrule
        w/o HP \& Conf.
        & 16.0 & 50.7 & 6.7
        & 98.7 & 43.0 \\
        w/o HP
        & 41.3 & \textbf{64.7} & 16.7
        & \textbf{99.3} & 55.5 \\
        PriDex
        & \textbf{98.0} & 44.7 & \textbf{40.0}
        & 46.0 & \textbf{57.2} \\
        \bottomrule
    \end{tabular*}
\end{table}

\subsubsection{Performance Analysis}
\label{sec:performance_analysis}
Table~\ref{tab:sim_rl_results} shows that PriDex outperforms the adapted H2S2R and ViViDex baselines on all four tasks.
The ablations further indicate task-dependent benefits from human-motion guidance and confidence adaptation, reflecting different trade-offs between action exploration and contact stability.

Human-motion guidance complements object references with additional cues about action direction.
Object trajectories describe the desired manipulation outcome and its evolution, but the policy must still discover robot actions that realize these motions.
Pouring requires a large tilting motion while maintaining a grasp.
Bottle-cap unscrewing requires one hand to stabilize the bottle while the other rotates the cap relative to it.
Our auxiliary reward uses arm-motion directions induced by wrist references to guide exploration of these coordinated actions, without requiring exact reproduction of full human-hand poses.
The success-rate drops on Pour and Unscrew after removing HP are consistent with directional guidance facilitating the exploration of effective coordinated actions.

In contrast, Clean and Sweep achieve higher success rates without HP.
Rollout inspection provides a task-specific explanation for Clean.
The source demonstration includes wiping along the plate rim,
and wrist-motion guidance encourages PriDex to retain this behavior.
Such motions can place greater demands on contact stability and
grasp maintenance.
Without HP, the policy tends to reduce rim-wiping motions and
adopts a simpler strategy that is less prone to failure while
still satisfying the cleaning success criterion.
These observations suggest that the improvement after removing HP
partly reflects greater freedom to simplify the demonstrated
behavior under the current evaluation criterion.

For Sweep, the improvement without HP may reflect greater
flexibility to adapt motion to the current contact state.
Together, these results indicate that the benefit of wrist-motion
guidance depends on both contact compatibility and alignment with
the evaluated task objective.
Although directional guidance can facilitate coordinated
exploration, it can also encourage demonstrated motions that
complicate stable execution.

Confidence adaptation adjusts object-tracking tolerances and
arm-residual bounds according to reference confidence.
High-confidence segments constrain tracking more tightly,
while low-confidence segments allow larger motion corrections.
With wrist guidance disabled, removing this adaptation reduces
mean success from $55.5\%$ to $43.0\%$.
Success decreases on Pour, Sweep, and Unscrew, while Clean
remains nearly unchanged.
This pattern supports adapting tracking constraints and
correction freedom to reference confidence during motion
refinement.

\subsection{From RL Rollouts to Visuomotor Policies}
We further evaluate whether PriDex-generated rollouts provide useful
supervision for visuomotor policy learning. For each task, we train
Diffusion Policies using successful rollouts from either full PriDex
or the w/o HP variant. Both datasets contain $200$ demonstrations per
source video ($600$ per task), and all other training settings are
matched. Each policy is evaluated over $150$ closed-loop trials per
task.

\begin{table}[t]
    \centering
    \caption{Distilled-policy success rates (\%).}
    \vspace{-3mm}
    \label{tab:distillation}
    \footnotesize
    \setlength{\tabcolsep}{3pt}
    \renewcommand{\arraystretch}{1.08}
    \begin{tabular*}{\linewidth}{@{\extracolsep{\fill}}lccccc@{}}
        \toprule
        Data source & Pour & Sweep & Unscrew & Clean & Mean \\
        \midrule
        w/o HP
        & 18.0 & 72.0 & 0.7
        & 6.0 & 24.2 \\
        PriDex
        & \textbf{24.0} & \textbf{80.0} & \textbf{5.3}
        & \textbf{16.0} & \textbf{31.3} \\
        \bottomrule
    \end{tabular*}
    \vspace{-5mm}
\end{table}

PriDex-generated demonstrations improve visuomotor success on Pour,
Sweep, Unscrew, and Clean, increasing the mean success rate from $24.2\%$ to
$31.3\%$. On Unscrew, success increases from $0.7\%$ to $5.3\%$, indicating that
this task remains challenging for visual-policy distillation under the
current training budget.

\subsection{Real-Robot Deployment Experiments}
We demonstrate real-world deployment on water pouring, sweeping into a dustpan, and plate cleaning using a Dexmate bimanual robot equipped with two Sharpa dexterous hands. Each policy receives RGB images from one head-mounted and two wrist-mounted cameras, together with proprioceptive observations comprising joint angles and wrist poses, and outputs absolute joint-position targets.

\noindent\textbf{Training Protocol.}
We compare two training conditions for each task. \emph{Real-only} trains a task-specific Diffusion Policy~\cite{chi2023diffusionpolicy} using $20$ successful real-world demonstrations collected through teleoperation. \emph{Sim+Real} first pretrains the policy on $120$ successful PriDex simulation demonstrations per task and then fine-tunes it on exactly the same $20$ real demonstrations. Fine-tuning uses only real-world data.

During deployment, the policy conditions on the two most recent observation steps and uses a prediction horizon of $16$ steps. It executes eight control steps before replanning from updated observations.

\noindent\textbf{Evaluation Protocol.}
Both training conditions are evaluated using the same objects and initial placements, without randomized perturbations. Pour succeeds when the bottle opening is aligned with the cup opening while the bottle is tilted beyond $90^{\circ}$. Sweep succeeds when the parts are swept into the dustpan. Clean succeeds when the cloth wipes back and forth over the plate four times. Returning the objects to their initial positions is not required. Real-robot trials were monitored by an operator without a fixed timeout. The operator terminated each trial upon task success, an object drop, or when the motion was judged unsafe or had deviated substantially such that successful completion was considered infeasible. Trials terminated before meeting the task-success criterion were recorded as failures.

\noindent\textbf{Real-Robot Results.}
We evaluated each training condition over $10$ trials per task. As shown in Table~\ref{tab:real_robot_results}, simulation pretraining increased mean success from $30.0\%$ to $63.3\%$ using the same real demonstrations, with improvements on all three tasks. These results show that PriDex provides useful pretraining data for bimanual dexterous manipulation with limited real demonstrations. Representative executions are shown in the supplementary video.

\begin{table}[htbp]
    \centering
    \caption{Real-robot results.}
    \label{tab:real_robot_results}
    \small
    \setlength{\tabcolsep}{5pt}
    \renewcommand{\arraystretch}{1.12}
    \begin{tabular*}{\linewidth}{@{\extracolsep{\fill}}lcc@{}}
        \toprule
        Task & Real-only & Sim+Real \\
        \midrule
        Pour  & $2/10$ & $\mathbf{5/10}$ \\
        Sweep & $3/10$ & $\mathbf{8/10}$ \\
        Clean & $4/10$ & $\mathbf{6/10}$ \\
        \midrule
        Mean success & $30.0\%$ & $\mathbf{63.3\%}$ \\
        \bottomrule
    \end{tabular*}
\end{table}

\section{CONCLUSION}

We presented PriDex, a framework that converts human-video priors into bimanual dexterous robot demonstrations through affordance-guided grasp initialization and confidence-guided RL refinement. PriDex achieves a $57.2\%$ mean simulation success rate and generates demonstrations that improve distilled visuomotor-policy performance on all four tasks under matched training budgets. The ablations highlight the task-dependent role of wrist-motion guidance and the value of adapting motion refinement to reconstruction confidence. Real-robot experiments further show that simulation pretraining improves policy success with the same limited real demonstrations. Future work will improve visual-policy distillation for challenging contact-rich tasks.

\textbf{Acknowledgment.} Generative AI tools were used to polish and refine the paper.

\bibliographystyle{IEEEtran}
\bibliography{root}

\end{document}